%% file: main.tex
\documentclass[10pt,twocolumn,letterpaper]{article}

\usepackage[pagenumbers]{cvpr} 
\usepackage{pifont}
\usepackage{multirow}
\newcommand{\workname}{Movie Weaver }

\input{preamble}


\definecolor{cvprblue}{rgb}{0.21,0.49,0.74}
\usepackage[pagebackref,breaklinks,colorlinks,allcolors=cvprblue]{hyperref}

\def\paperID{5034} 
\def\confName{CVPR}
\def\confYear{2025}

\begin{document}

\twocolumn[{

\title{Movie Weaver: Tuning-Free Multi-Concept Video Personalization with Anchored Prompts
\vspace{-1em}
}


\author{Feng Liang\footnotemark[1]~~\textsuperscript{\rm 1},
        Haoyu Ma\textsuperscript{\rm 2}, 
        Zecheng He\textsuperscript{\rm 2},
        Tingbo Hou\textsuperscript{\rm 2},
        Ji Hou\textsuperscript{\rm 2},
        Kunpeng Li\textsuperscript{\rm 2}, 
        Xiaoliang Dai\textsuperscript{\rm 2}, \\
        Felix Juefei-Xu\textsuperscript{\rm 2}, 
        Samaneh Azadi\textsuperscript{\rm 2},
        Animesh Sinha\textsuperscript{\rm 2},
        Peizhao Zhang\textsuperscript{\rm 2}, 
        Peter Vajda\textsuperscript{\rm 2}, 
        Diana Marculescu\textsuperscript{\rm 1}\\
\normalsize{\textsuperscript{\rm 1}The University of Texas at Austin, Chandra Family Department of Electrical and Computer Engineering \textsuperscript{\rm 2}Meta GenAI}\\
\texttt{\{jeffliang,dianam\}@utexas.edu}, \texttt{\{haoyuma,zechengh,stzpz\}@meta.com}\\
\texttt{\href{https://jeff-liangf.github.io/projects/movieweaver}{https://jeff-liangf.github.io/projects/movieweaver}}
}

\maketitle

\vspace{-3em}

\begin{center}
    \centering
    \includegraphics[width=0.8\textwidth]{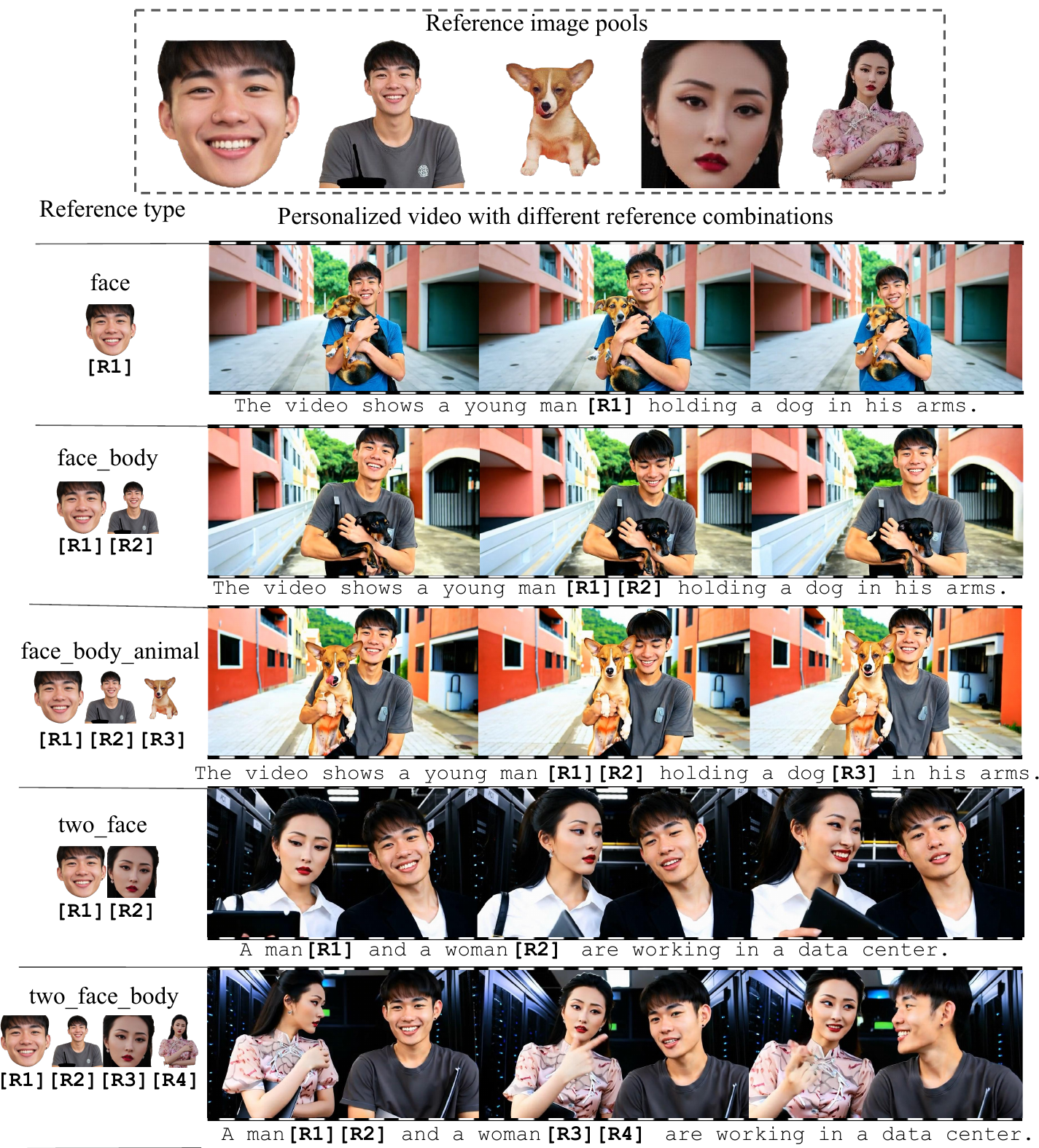}
    \vspace{-0.5em}
    \captionof{figure}{
    We introduce \textbf{Movie Weaver}, a video diffusion model for personalized multi-concept video creation. 
    Besides text prompts, our model allows users to input different combinations of reference images, \eg, face, body, and animal images, to customize videos in a tuning-free manner. The left column displays different types of reference images, while the right column shows the generated videos, with \emph{anchored prompt} listed beneath each video. We encourage readers to check our video results in the supplementary materials.
    }
    \label{fig:teaser}
\end{center}
}]

\footnotetext[1]{Work partially done during an internship at Meta GenAI.}
\input{sec/0_abstract}    
\input{sec/1_intro}
\input{sec/2_related_work}
\input{sec/3_method}
\input{sec/4_experiment}
\input{sec/5_conclusion}

{
    \small
    \bibliographystyle{ieeenat_fullname}
    \bibliography{main}
}

\input{sec/X_suppl}

\end{document}

%% file: preamble.tex
\definecolor{americanrose}{rgb}{1.0, 0.01, 0.24}

%% file: sec/0_abstract.tex
\begin{abstract}

Video personalization, which generates customized videos using reference images, has gained significant attention.
However, prior methods typically focus on single-concept personalization, limiting broader applications that require multi-concept integration.
Attempts to extend these models to multiple concepts often lead to identity blending, which results in composite characters with fused attributes from multiple sources.
This challenge arises due to the lack of a mechanism to link each concept with its specific reference image.
We address this with anchored prompts, which embed image anchors as unique tokens within text prompts, guiding accurate referencing during generation.
Additionally, we introduce concept embeddings to encode the order of reference images.
Our approach, Movie Weaver, seamlessly weaves multiple concepts—including face, body, and animal images—into one video, allowing flexible combinations in a single model.
The evaluation shows that \workname outperforms existing methods for multi-concept video personalization in identity preservation and overall quality.

\end{abstract} 
\vspace{-1em}

%% file: sec/1_intro.tex
\section{Introduction}
\label{sec:intro}

Foundational text-to-video generation models~\cite{ho2022imagen,villegas2022phenaki,singer2022make,zhou2022magicvideo,chen2023videocrafter1,blattmann2023align,girdhar2023emu,guo2023animatediff,kondratyuk2023videopoet,bar2024lumiere,yang2024cogvideox,chen2024gentron,sora,luma,gen3,kling,genmo2024mochi} have made substantial progress in the past few years.
Leveraging these advancements, personalized video generation enables users to create customized videos with their images, offering huge potential for applications like consistent storytelling.
However, prior efforts~\cite{he2024id, wu2024customcrafter, moviegen} primarily support single concept (face or object) personalization, limiting their use in complex, real-world scenarios.
Practical applications often require multi-concept compositions, like interactions between two people or between a person and a pet.
To meet this need, we introduce Movie Weaver, a video diffusion model that \emph{weaves} diverse image combinations and text prompts to create personalized multi-concept videos, as illustrated in Figure~\ref{fig:teaser}.

Established video personalization methods~\cite{he2024id, wu2024customcrafter, jiang2024videobooth, moviegen} typically extract vision tokens via an image encoder~\cite{radford2021learning}, then inject these tokens into diffusion models through cross attention.
To support multiple face references, we extend this approach by directly feeding concatenated vision tokens from multiple references into the cross attention layer.
While this direct extension works in some cases, it often encounters a severe identity blending issue~\cite{kumari2023multi,xiao2024fastcomposer}, generating composite characters that fuse attributes from both references.
The issue is more severe when the character faces are close, \ie, from the same gender or race. 
This is because such a direct extension lacks the ability to associate each reference image with their descriptions within the prompt.
When vision tokens from different faces are similar, the model struggles to differentiate between them, resulting in identity blending.

To address this issue, we explicitly build the association between the concept description and the corresponding reference image. 
We introduce \emph{anchored prompts} by injecting unique text tokens (\texttt{[R1]}) after each concept description, as shown in Figure~\ref{fig:teaser}.
Upon encountering \texttt{[R1]}, the model links it with the matching reference image of \texttt{[R1]} and uses it as the visual supplement for the concept.
This anchored prompt approach extends easily to multiple concepts by adding more anchors (\eg, \texttt{[R2]}, \texttt{[R3]}) and corresponding images.
Notably, anchored prompts only require input modifications without any architectural changes like identity-specific tuning~\cite{kumari2023multi,kwon2024concept}, predefined layout~\cite{liu2023cones,gu2024mix} or masked attention~\cite{xiao2024fastcomposer,kim2024instantfamily,he2024uniportrait,ostashev2024moa}. 
This preserves architectural simplicity and leverages scalability.

With the anchored prompts established, the next question is how to distinguish the reference images of \texttt{[R1]} and \texttt{[R2]}.
Our baseline architecture~\cite{he2024id,jiang2024videobooth,moviegen} concatenates the vision tokens with text tokens and feeds them together to the cross attention layer of diffusion models.
However, since cross attention is order-agnostic, we must inject some information about the order of reference images.
Inspired by positional encoding~\cite{vaswani2017attention,dosovitskiy2020image}, we introduce \emph{concept embedding}, adding a unique embedding to each set of vision tokens from the reference image.
This is different from traditional positional encoding, where different embeddings are assigned to individual tokens. Our method applies a uniform concept embedding to all tokens from the same image.

We also propose an automatic data curation pipeline to get anchored prompts and ordered reference images.
This pipeline supports diverse reference combinations (such as face, face-body, face-body-animal, two-face, and two-face-body) by leveraging a suite of foundations models~\cite{liu2023grounding,kirillov2023segment,radford2021learning,dubey2024llama}, yielding a dataset of 230K videos. 
Using the proposed anchored prompts and concept embeddings, we continue training \workname on a pre-trained single-face video personalization model~\cite{moviegen}. 
As showcased in Figure~\ref{fig:teaser}, our model effectively generates high-quality videos with diverse reference combinations without additional tuning. 
Compared with proprietary Vidu 1.5~\cite{vidustudio} and extended baseline, \workname exceeds in identity preservation and visual quality.

Our contributions are three-fold: (1) \textbf{Anchored Prompts:} We introduce anchored prompts to link specific reference images with concept description, resolving identity blending in multi-concept video personalization without architectural changes. (2) \textbf{Concept Embeddings:} We use unique embeddings for each reference image to maintain identity and order in multi-reference settings. (3) \textbf{Automatic Data Curation:} We implement a pipeline to curate training data with diverse reference combinations, enabling high-quality, tuning-free multi-concept video generation.

%% file: sec/2_related_work.tex
\section{Related Work}
\label{sec:related_work}

\paragraph{Personalized image generation}
Personalized generation begins with identity-specific tuning methods that further finetune a text-to-image model on a set of reference images. 
For instance, Textual Inversion~\cite{gal2022image} finetunes special text tokens for the target identity. 
DreamBooth~\cite{ruiz2023dreambooth} further conducts end-to-end model finetuning besides tuning the special text tokens.
Custom Diffusion~\cite{kumari2023multi} extends a parameter-efficient finetuning technique to incorporate multiple concepts. 
However, these tuning-based methods require separate optimizations for every concept, which does not scale well in real applications.
Recent tuning-free methods train one base personalization model and then use it for arbitrary reference images in inference.
For example, ELITE~\cite{wei2023elite}, PhotoMaker~\cite{li2023photomaker}, PhotoVerse~\cite{chen2023photoverse}, IP-Adapter~\cite{ye2023ip} , InstantID~\cite{wang2024instantid}, and Imagine Yourself~\cite{dai2023emu,meta24memu} all leverage a vision encoder to extract visual tokens from the reference image and inject them to the diffusion process.
Our method \workname falls in the second tuning-free method which targets multi-concept personalization.

\paragraph{Personalized video generation}
While personalized image generation has shown promising results, personalized video generation remains an unsolved problem.
Compared to static images, personalized videos require more diverse and complex modifications on the reference image, \eg, turning the head, changing poses, and camera motion movements.
Pioneering work VideoBooth~\cite{jiang2024videobooth}, DisenStudio~\cite{chen2024disenstudio}, Magic-Me~\cite{ma2024magic}, DreamVideo~\cite{wei2024dreamvideo}, CustomVideo~\cite{wang2024customvideo}, TweedieMix~\cite{kwon2024tweediemix}, MultiConceptVideo~\cite{kothandaraman2024text} and CustomCrafter~\cite{wu2024customcrafter} use identity-specific finetuning to inject identity into a video generation model. 
However, these methods require separate fine-tuning for every identity, which limits their applications.
Another line of work includes tuning-free methods.
ID-Animator~\cite{he2024id}, MovieGen~\cite{moviegen} and Video Alchemist~\cite{chen2025multi} train a vision encoder~\cite{radford2021learning} to inject the reference image features into the diffusion models.
Our Movie Weaver falls in the tuning-free setting with a special focus on multi-concept personalization.

\paragraph{Multi-concept personalization}
Multi-concept personalization aims to generate harmonized content with multiple reference concepts.
Identity-specific tuning methods, \eg, Custom Diffusion~\cite{kumari2023multi}, Break-A-Scene~\cite{avrahami2023break}, Concept Weaver~\cite{kwon2024concept} and MuDI~\cite{jang2024identity}, achieve multi-concept personalization by finetuning the multiple text embeddings and model weights.
Tuning-free methods, \eg, FastComposer~\cite{xiao2024fastcomposer}, MoA~\cite{ostashev2024moa}, InstantFamily~\cite{kim2024instantfamily} and UniPortrait~\cite{he2024uniportrait}, train on large-scale text-image datasets and inject multiple image embeddings directly into diffusion process during inference.
Mix-of-show~\cite{gu2024mix} and OMG~\cite{kong2024omg} merge single-concept models but are limited by the availability of community models, primarily covering well-known intellectual property (IP) or celebrities.
Unlike these methods that separate concepts using predefined layout~\cite{liu2023cones,gu2024mix} or masked attention~\cite{xiao2024fastcomposer,kim2024instantfamily,kim2024instantfamily,ostashev2024moa} or word/image feature concatenation~\cite{chen2025multi}, our \workname proposes anchored prompts and concept embeddings to link concept with matched reference image without architectural change.
Concurrent work ViMi~\cite{fang2024vimi} is the closest to our setting but requires complex data retrieval to curate data and an additional multimodal LLM~\cite{liu2024visual} for the image-text process. In contrast, our data curation pipeline operates without supplementary retrieval data, while maintaining simplicity in architecture.

%% file: sec/3_method.tex
\section{Challenges in Extending Single-Concept to Multi-Concept Personalization}

\begin{figure}[t]
    \begin{center}
    \includegraphics[width=0.9\linewidth]{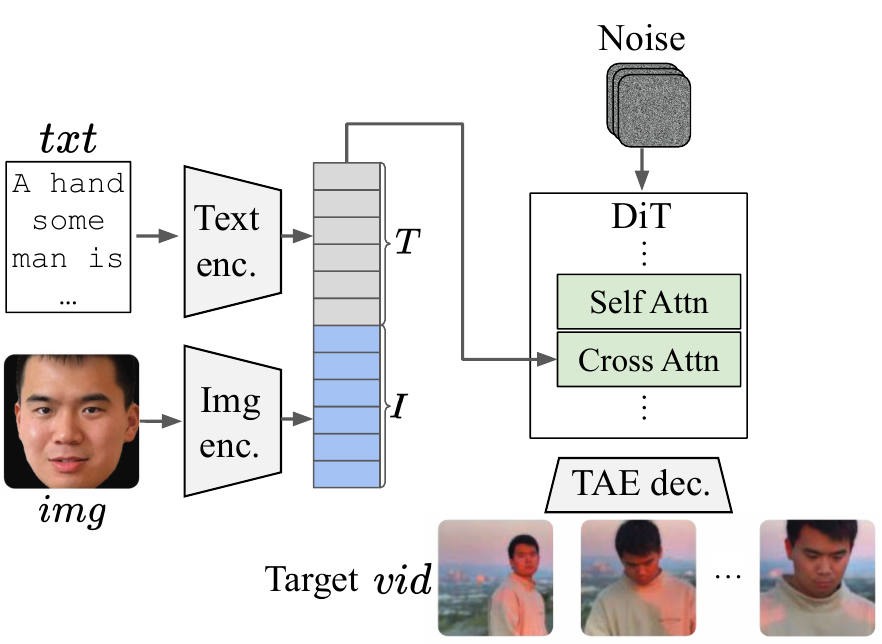}
    \end{center}
    \vspace{-1.5em}
        \caption{\textbf{Single-concept personalization architecture.} Building on a pre-trained text-to-video model, this approach adds an image encoder to process reference images. Image and text tokens are concatenated and fed into cross attention layer.}
        \vspace{-1em}
    \label{fig:single_concept_arch}
\end{figure}

\subsection{Singe-concept video personalization}

Based on a pre-trained text-to-video diffusion model, prior single-concept video personalization methods~\cite{he2024id, wu2024customcrafter, jiang2024videobooth, moviegen} typically introduce an additional CLIP image encoder~\cite{radford2021learning,xu2023demystifying} to process reference images. 
As shown in Figure~\ref{fig:single_concept_arch}, given a dataset triplet $\{txt, img, vid\}$—representing text prompt, reference image, and target video—these methods extract text tokens $T$ and image tokens $I$ via text and image encoders. 
The tokens $I$ and $T$ are concatenated and then fed into the cross attention layer of the diffusion transformer~\cite{peebles2023scalable}.

\vspace{-1em}
\begin{equation}
\text{CrossAttn} = \text{softmax}\left(\frac{Q_Z\left[ K_T, K_I \right]^{Tr}}{\sqrt{d}}\right) \left[V_T, V_I \right]
\label{eq:cross_attention}
\end{equation}

Here, the query is derived from the video latent $Z$, and the key/value is derived from the concatenation of $I$ and $T$. 
$\left[ \cdot \right]$ and $^{Tr}$ denotes the concatenation and transpose operation, respectively. 
Key and value are derived through linear projection, so $\left[V_T, V_I \right] = V_{\left[T,I\right]}$. For simplicity, we maintain formal notation without specifying this further.

\subsection{Naive extension to multi-concept}

\begin{figure}[t]
    \begin{center}
    \includegraphics[width=\linewidth]{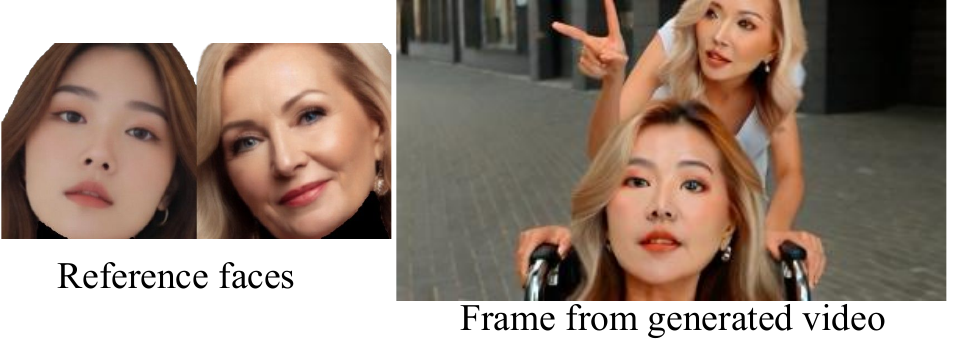}
    \end{center}
    \vspace{-2em}
        \caption{\textbf{Identity blending} generates composite faces with characteristics from both references. Text prompt: \texttt{"A woman in wheelchair discussing with a woman nurse."}}
        \vspace{-1em}
    \label{fig:identity_blending}
\end{figure}

To support multi-concept personalization, a straightforward approach is preparing image tokens from multiple reference images. 
We began by experimenting with two-face personalization. 
According to Equation~\ref{eq:cross_attention}, this results in $ K_I = \left[ K_{I_1}, K_{I_2} \right]$ and $ V_I = \left[ V_{I_1}, V_{I_2} \right]$, where $I_1$ and $I_2$ represents tokens of two different faces.
This naive approach, however, led to severe identity blending~\cite{kumari2023multi,xiao2024fastcomposer}, where the characteristics of two faces would fuse together, resulting in a composite one as seen in Figure~\ref{fig:identity_blending}. 
The issue arises because the model cannot effectively link each concept description to its corresponding image. 
In cross attention, a query latent in $Q_Z$ can attend to the entire prompt, including two text tokens of "\texttt{woman}", and cannot distinguish between vision tokens $I_1$ and $I_2$ due to order-agnostic processing. 
For accurate video generation, the model must link "\texttt{A woman in wheelchair}" to the first image and "\texttt{a woman nurse}" to the second image.

\section{Movie Weaver}
\label{sec:method}

Unlike traditional approaches that require concept-specific tuning or architectural modifications, Movie Weaver aims to enable multi-concept video personalization with architectural simplicity and flexibility. 
Movie Weaver introduces two novel components, namely anchored prompts and concept embeddings, alongside an automatic data curation pipeline, all of which allow accurate, tuning-free multi-concept video generation.

\begin{figure*}[t]
    \centering
    \vspace{-2em}
    \includegraphics[width=0.98\textwidth]{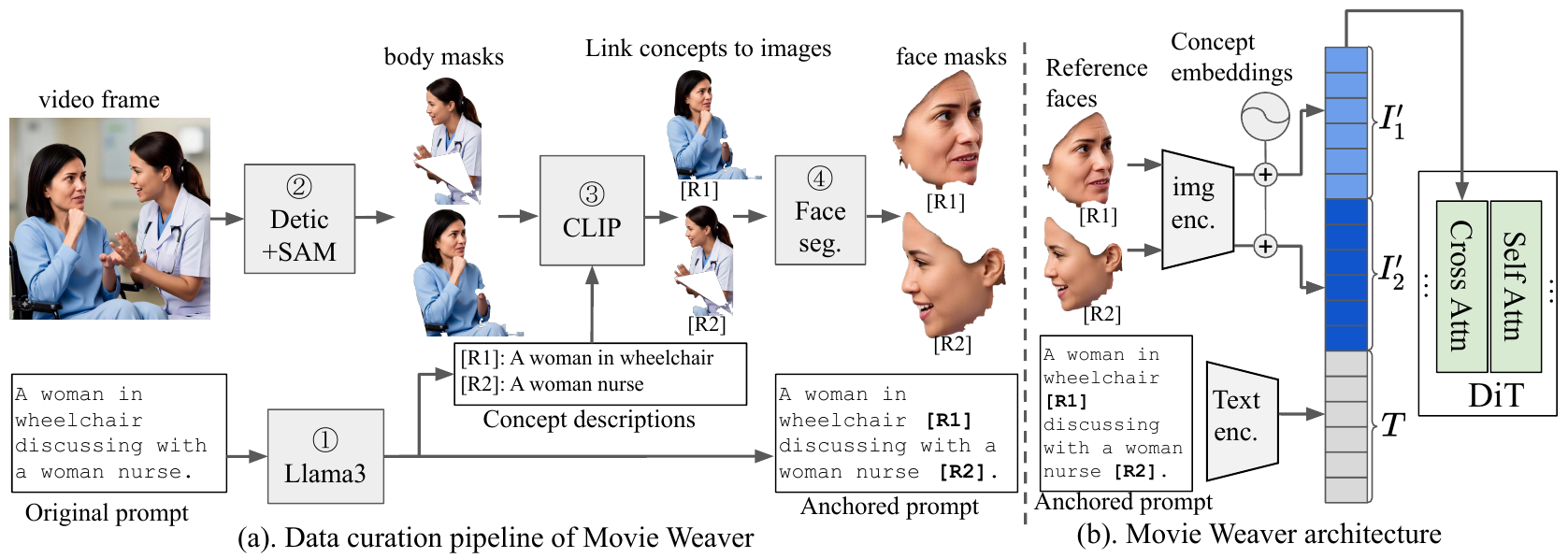}
    \vspace{-0.5em}
    \caption{(a) \textbf{Data curation.} For a video-text pair, \ding{172} concept descriptions and anchored prompts are generated via in-context learning with Llama-3. After \ding{173} extracting body masks, \ding{174} CLIP links each concept to its corresponding image. \ding{175} Finally, face images are obtained using a face segmentation model. (b) \textbf{\workname architecture.} Compared to the single-concept baseline, reference images are arranged in a specific order for concept embedding, and anchored prompts are utilized. Shared components are omitted for simplicity.}
    \label{fig:data_curation}
    \vspace{-1em}
\end{figure*}

\subsection{Anchored prompts}

The success of multi-concept personalization lies in accurately associating each concept with its corresponding image. 
Previous methods often rely on predefined layouts~\cite{liu2023cones,gu2024mix} or complex masked attention modules~\cite{xiao2024fastcomposer,kim2024instantfamily,he2024uniportrait,ostashev2024moa} to establish these associations, which increases model complexity and limits flexibility.
Our \workname introduces a streamlined solution with anchored prompts.
For prompt in Figure~\ref{fig:identity_blending}, we use Llama-3~\cite{dubey2024llama} to identify the concept descriptions "\texttt{A woman in wheelchair}" and "\texttt{a woman nurse}", we then append unique tokens (\eg, \texttt{[R1]}, \texttt{[R2]}) after each description, creating the prompt "\texttt{A woman in wheelchair [R1] discussing with a woman nurse [R2].}"
Ordered reference images are then linked to these unique tokens, allowing the model to associate each description precisely with its reference image.
This approach offers two key advantages: (1) Flexibility: anchored prompts can easily extend to different descriptions, such as body or animal descriptions. Moreover, it allows to append multiple references, such as face and body images, on the same person.; and (2) Simplicity: this approach requires only input modifications, allowing Movie Weaver to retain an architecture similar to single-concept models.

\subsection{Concept embeddings}

While anchored prompts establish explicit associations, we also need to encode the order information of reference images. 
The cross attention mechanism, as outlined in Equation~\ref{eq:cross_attention}, is inherently order-agnostic, \ie, swapping the order of two reference images yields identical results.
To address this, Movie Weaver introduces concept embeddings, a novel adaptation of positional encoding~\cite{vaswani2017attention} tailored to multi-concept personalization. 
Specifically, given $I_1$ and $I_2$ as image tokens from two different reference images, we add the same concept embedding $Pos_1$ and $Pos_2$  to each set of tokens, respectively:

\begin{equation}
I'_1 = I_1 + Pos_1,  I'_2 = I_2 + Pos_2, ...
\label{eq:concept_embedding}
\end{equation}

Here, $I_1$ and $I_2$ have dimensions $[N, C]$, while $Pos_1$ and $Pos_2$ are of shape $[1, C]$, where $N$ is the number of tokens and $C$ is the feature dimension. Using the broadcasting of Pytorch, the same concept embedding is added to the entire set. 
This is different from traditional positional encoding where distinct embeddings are added to individual tokens.
We also experimented with per-token positional encoding, but it produced less effective results compared to using a concept embedding for each set of vision tokens (see Section~\ref{sec:ablation_ap_ce}). 

\subsection{Data curation pipeline}
\label{sec:overall_architecture}

Starting with text-video pairs, we curate data by leveraging a set of foundation models.
We take the preparation of two-face configuration as an example in Figure~\ref{fig:data_curation} (a). 
We first use in-context learning of Llama-3~\cite{dubey2024llama} to extract concept descriptions from the original prompt and get a rewritten anchored prompt. 
For the reference frame, typically the first frame of the video clip, we use a detection model Detic~\cite{zhou2022detecting} and a segmentation model SAM~\cite{kirillov2023segment} to extract the subject masks.
Using the detection results, we can also tell the number of people in the video and other objects in the video, which helps us filter the data.
Then, a pre-trained CLIP~\cite{radford2021learning} assigns the concept descriptions to the body images, establishing the link between \texttt{[R]} and reference images. Lastly, we use a face segmentation model to extract face masks from body images.
While this example only illustrates the two-face scenarios, the approach can be naturally extended to other combinations with more \texttt{[R]}s and reference images. 
More examples can be found in the supplementary materials. 

With ordered reference images and rewritten anchored prompts prepared, our Movie Weaver architecture is illustrated in Figure~\ref{fig:data_curation}(b). 
Compared with single-concept architecture in Figure~\ref{fig:single_concept_arch}, our model requires reference images in a specific order to apply concept embeddings effectively. We also need to have corresponding anchored prompts to strengthen the results.

%% file: sec/4_experiment.tex
\section{Experiments}
\label{sec:experiments}

\subsection{Implementation details}

\begin{figure*}[t]
    \centering
    \vspace{-1em}
    \includegraphics[width=0.95\textwidth]{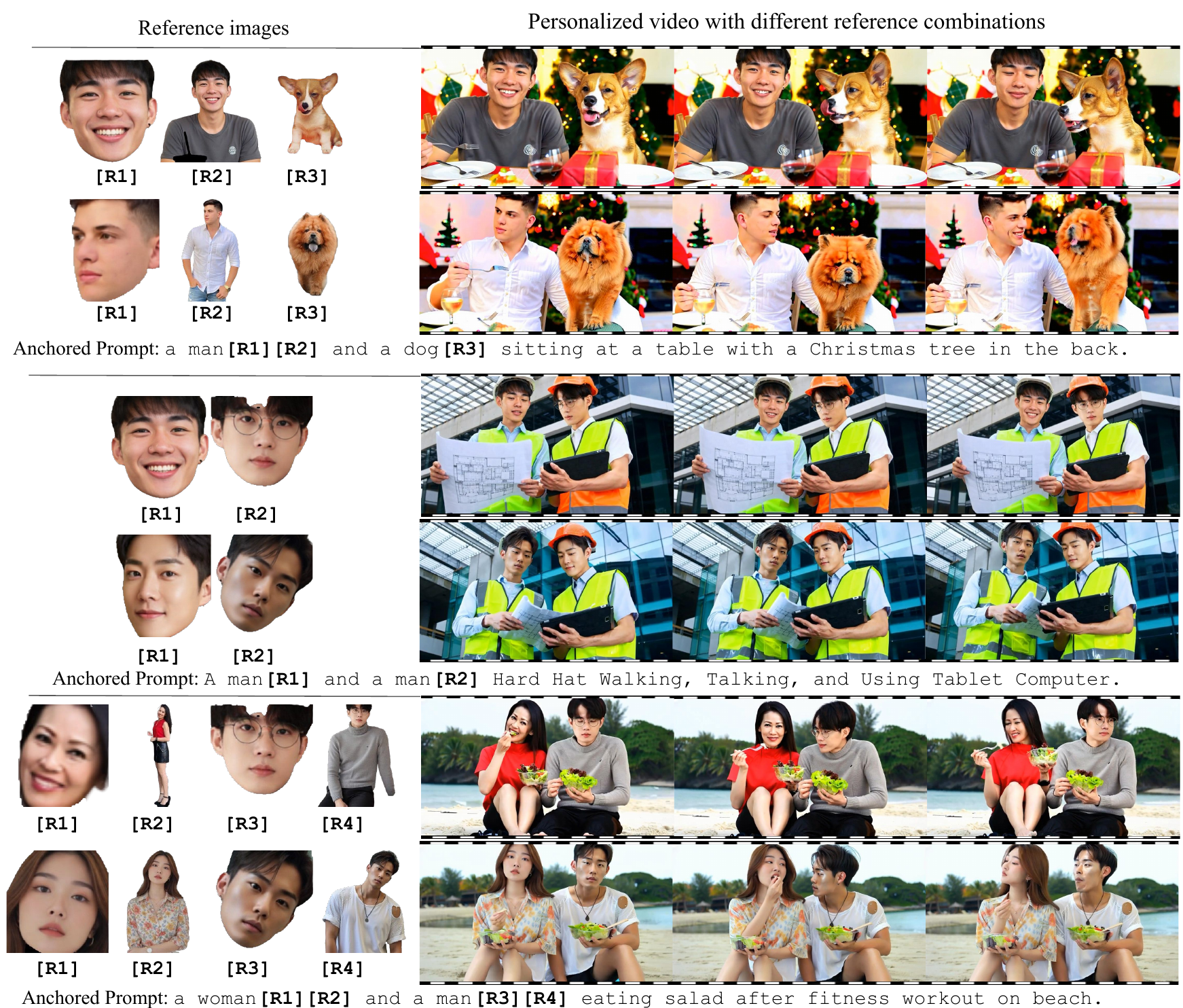}
    \vspace{-0.5em}
    \caption{\textbf{Qualitative results of Movie Weaver.} \workname supports different combinations of reference images and can generate high-quality videos with high identity preservation. We encourage readers to check our video results in the supplementary materials.}
    \label{fig:mpt2v_results}
    \vspace{-0.5em}
\end{figure*}

\textbf{Pretraining data} Our Movie Weaver supports 5 configurations: face, face-body, face-body-animal, two-face and two-face-body, as showcase in Figure~\ref{fig:teaser}.
We collect all our videos from Shutterstock Video~\cite{ShutterstockVideo}.
For the face and face-body configurations, we curated 100K videos featuring only a single person performing various activities.
In the two-face and two-face-body configurations, videos were selected based on the presence of exactly two individuals, verified via a detection model, resulting in a dataset comprising 118K videos. 
For the face-body-animal setup, we found it more challenging to source videos featuring both a person and an animal, ultimately assembling a collection of 10K videos.
We conduct mix-pretraining, where we equally sample examples for each configuration using data resampling. More information can be found in Section~\ref{sec:mix_training}.

\textbf{Finetuning data} Existing research~\cite{dai2023emu,he2024imagine,moviegen} suggests that further finetuning on a small-scale, very high quality data significantly enhances visual quality and subject motion. 
We follow this principle by manually selecting videos with large human motion, rich human iterations, and high visual aesthetic. 
Same as pretraining data, we source different videos for our different configurations.
In summary, our finetuning set has 291 videos for face and face-body, 175 videos for two-face and two-face-body, and 185 videos for face-body-animal.

\textbf{Model details}
We adopt the Movie Gen architecture~\cite{moviegen}, with two versions: a 4B parameter model for the ablation study and a 30B parameter model for the final results. 
Unless specified otherwise, Movie Weaver refers to the 30B model. 
Movie Weaver uses MetaCLIP~\cite{xu2023demystifying} as image encoder and uses three text encoders: MetaCLIP~\cite{xu2023demystifying}, ByT5~\cite{xue2022byt5} and UL2~\cite{tay2022ul2}. 
We implement concept embeddings (CE) using learnable nn.Embedding in PyTorch, applied only to image embeddings.
The diffusion model contains 48 layers of diffusion transformers~\cite{peebles2023scalable}.
The temporal VAE has a compression rate of $8\times8\times8$, represents $8\times$ dimension reduce for spatial height/width and temporal frame.
Using an additional $2\times2\times1$ patchification, we compressed each 128-frame landscape video with 544$\times$960 resolution into a token sequence of length 32K.

\textbf{Training hyperparameters}
We initialize \workname from a pre-trained single-face personalization Movie Gen checkpoint. 
The model is trained with a learning rate of 1e-5 using the AdamW~\cite{loshchilov2017decoupled} optimizer for 20K iterations with a batch size of 32.
The training objective is flow matching~\cite{lipman2022flow} with optimal transport path. 
It took around 5 days to do pretraining on a cluster of 256 H100 GPUs.
Following this, we performed supervised finetuning with a reduced learning rate of 2e-6 for an additional 2K iterations.

\textbf{Test data}
Our test set has 5 configurations with 300+ reference-prompt pairs each, covering diverse ethics, human genders, animal types, and prompt styles. Human images are generated via text-to-image models EMU~\cite{dai2023emu}, while animal images are from external datasets (not in our training) like DreamBooth~\cite{ruiz2023dreambooth}. Prompts are generated via LLM in-context learning.

\subsection{Results}

\begin{figure*}[t]
    \centering
    \includegraphics[width=0.95\textwidth]{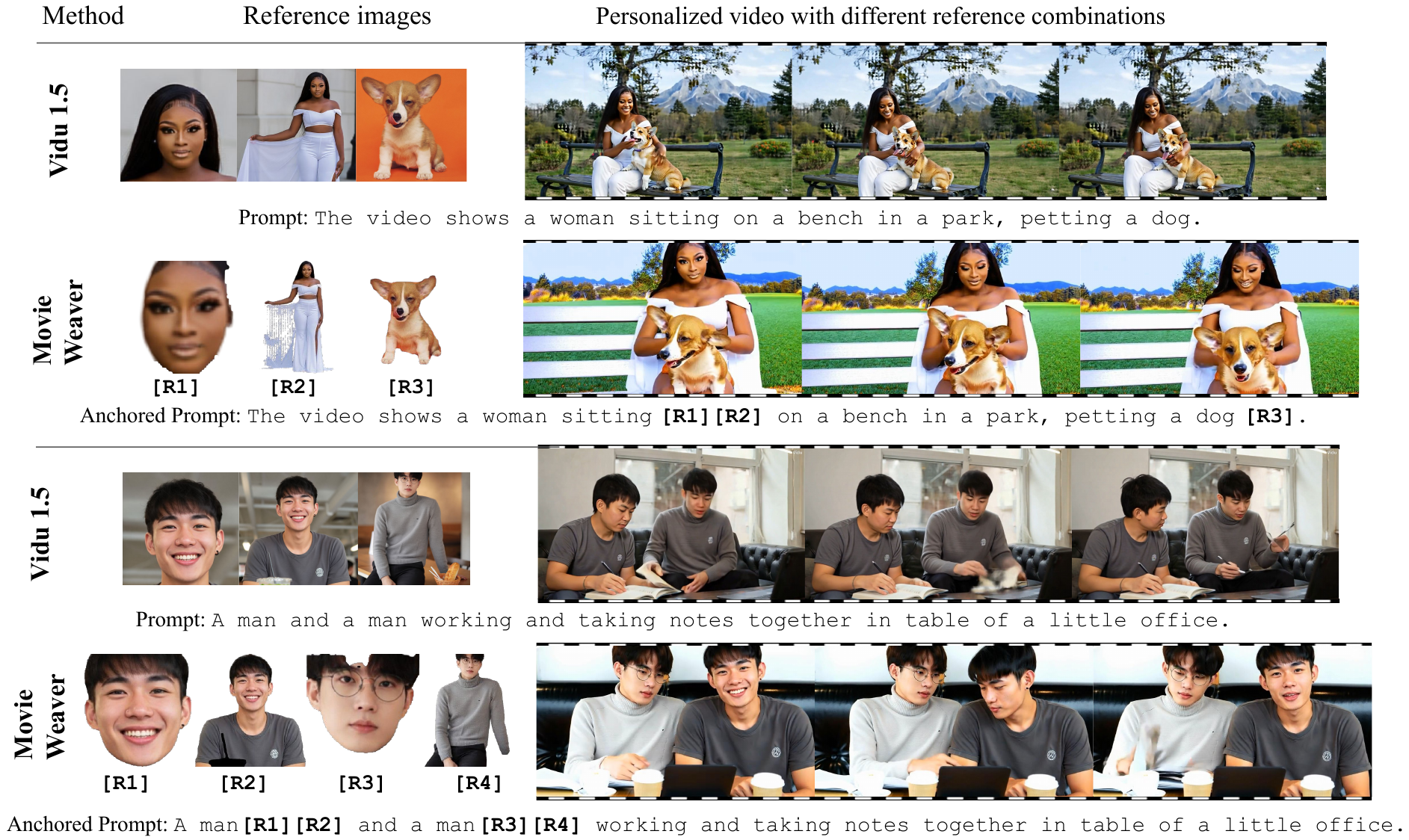}
    \vspace{-0.5em}
    \caption{\textbf{Comparison with state-of-the-art multi-concept video methods.} Compared to proprietary Vidu 1.5, Movie Weaver demonstrates superior identity preservation for both human and animal reference images.}
    \label{fig:comparison_sota}
    \vspace{-1em}
\end{figure*}

\subsubsection{Performance highlight}
We demonstrate three configurations of Movie Weaver: face-body-animal, two-face, and two-face-body, in Figure~\ref{fig:mpt2v_results}. 
For the same text prompt, we generate two different videos with two different sets of reference images. 
Notably, no facial, clothing, or dog descriptions are included in the prompts; all identity information is derived solely from the reference images. 
We highlight key features of our Movie Weaver: 
(1) Identity Preservation: Face, body, and animal details are accurately maintained in the generated videos.
Even small clothing details, such as the logo on the gray T-shirt in the first video and the tear on the white T-shirt in the sixth, are faithfully preserved.
In challenging two-face scenarios with same-gender, same-race pairs, Movie Weaver effectively retains each identity.
(2) Flexibility with References: Movie Weaver can adapt reference images to match the prompt without having to strictly follow. 
For instance, in the second video, the reference body image shows a man standing, yet the prompt requires him to sit. 
Our \workname selectively uses the upper body to align with the prompt.
Also for the fifth video, the standing woman body reference image is adapted to appear seated on a beach.
(3) Rich iteration between subjects: Beyond preserving identity, the generated videos capture dynamic interactions between subjects. In the first and second videos, the person interacts well with the dog, while in third through sixth videos, the two people display rich engagement.

\subsubsection{Comparison with existing methods}

\textbf{Multi-concept video personalization.} We compare \workname with the proprietary Vidu 1.5~\cite{vidustudio}, which is the state-of-the-art model that supports multi-concept video personalization. 
We observed their demos use cropped images, so we prepared the reference images without masking.
In the first face-body-animal example of Figure~\ref{fig:comparison_sota}, \workname shows better identity preservation for both the African American woman and the dog. 
While Vidu 1.5 correctly identifies the dog as a Corgi, it fails to reconstruct the distinctive white spots on the dog's face, which are crucial to the dog's identity.
In the second two-face-body example, because Vidu 1.5 supports a maximum of 3 reference images, we only input the face of the first character. 
Vidu 1.5 suffers from severe identity blending issues with the two generated characters looking and wearing similarly to each other, whereas \workname maintains clear distinctions between the two individuals.

\begin{figure}[t]
    \begin{center}
    \includegraphics[width=0.85\linewidth]{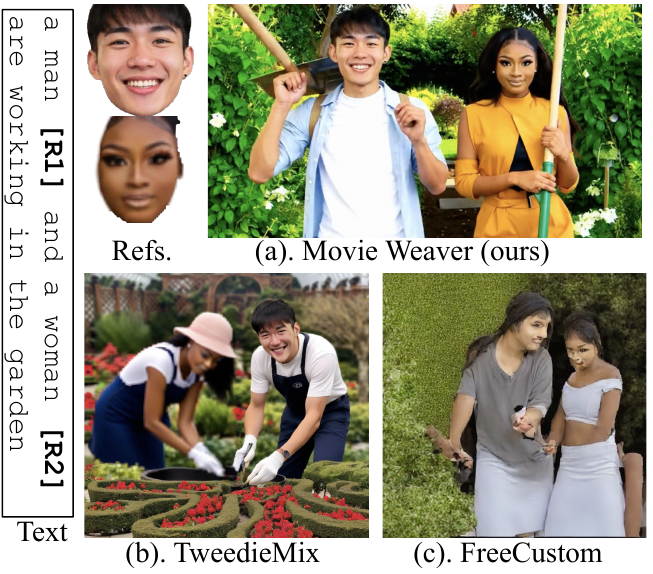}
    \end{center}
    \vspace{-2em}
        \caption{\textbf{Comparison with multi-concept image methods.} Movie Weaver has a better identity preserving and visual quality.}
        \vspace{-1em}
    \label{fig:multi_concept_t2i}
\end{figure}

\textbf{Multi-concept image personalization.} 
We also compare with representative multi-concept image personalization methods. 
For Tweediemix~\cite{kwon2024tweediemix}, we first fine-tune the base SDXL~\cite{podell2023sdxl} model for each reference concept using LORA~\cite{hu2021lora}, then conduct multi-concpet sampling using Tweedie’s formula. 
Because Tweediemix requires background reference, we select one of its pretrained garden LORA weights.
Freecustom~\cite{ding2024freecustom} is a tuning-free method, so we follow its practice by preparing two reference faces.
We select the first frame of our Movie Weaver to compare with these image methods.
As shown in Figure~\ref{fig:multi_concept_t2i}, our Movie Weaver preserves a much better identity and has higher visual quality when compared with TweedieMix and FreeCustom.

\textbf{Single-concept video personalization.} Although our Movie Weaver targets at multi-concept, it can perform well on single-concept setting. We compare with ID-Animator~\cite{he2024id} and DreamVideo~\cite{wei2024dreamvideo}, using 97 single-face personalization data. Following DreamVideo, we use four metrics, CLIP-T, CLIP-I, DINO-I, and Temporal Consistency. 
As we can see in Table~\ref{tab:quant_comp_singleface}, our Movie Weaver excels in all the metrics.

\begin{table}[t]
\vspace{-0.8em}
\centering
\small
\setlength{\tabcolsep}{3pt} 
\begin{tabular}{lcccc}
\toprule
\textbf{Method} & \textbf{CLIP-T} & \textbf{CLIP-I} & \textbf{DINO-I} & \textbf{Tem. Cons.} \\
\midrule
DreamVideo  & 0.282     &  0.498      &  0.246      &  0.956  \\
ID-Animator & 0.274     &  0.642      &  0.405      &  0.985  \\
Movie Weaver (ours) & \textbf{0.293}  &  \textbf{0.659}      &  \textbf{0.421}      &  \textbf{0.997}  \\
\bottomrule
\end{tabular}
\vspace{-1em}
\caption{\textbf{Quantitative comparison of single-face video personalization.} Movie Weaver excels in four metrics.}
\label{tab:quant_comp_singleface}
\end{table}

\begin{figure}[t]
    \centering
    \includegraphics[width=0.43\textwidth]{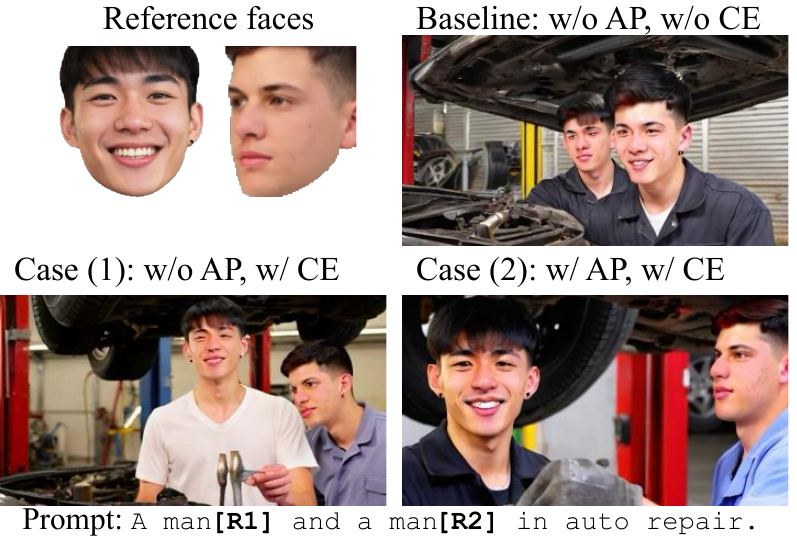}
    \small
    \begin{tabular}{ccccccc}
        \toprule
    \multirow{2}{*}{Case} & \multicolumn{2}{c}{Modules} & \multicolumn{3}{c}{Human study metrics} \\
    \cmidrule(lr){2-3} \cmidrule(lr){4-6}
    & AP & CE & sep\_yes$\uparrow$ & face1\_sim$\uparrow$ & face2\_sim$\uparrow$ \\
        \midrule
        Baseline &    &    & 42.9 & 3.4 & 3.0 \\
        (1)        &    & \checkmark & 98.2 & 58.8  & 41.9  \\
        (2)        & \checkmark & \checkmark & 99.3 & 66.8 & 66.1  \\
        \bottomrule
    \end{tabular}
    \vspace{-1em}
    \caption{\textbf{Ablation study of Anchored Prompts (AP) and Concept Embeddings (CE).} The top part shows the effect of AP and CE, while the bottom presents results from a human study. Metric sep\_yes indicates the percentage of cases where the two generated faces are distinguishable (\ie, no identity blending), face1\_sim and face2\_sim represent where a similar face to the left or right reference face, respectively, is found in the generated video.}
    \vspace{-1em}
\label{fig:ablation_ap_ce}
\end{figure}

\begin{figure*}[t]
    \centering
    \vspace{-1.55em}
    \includegraphics[width=0.93\textwidth]{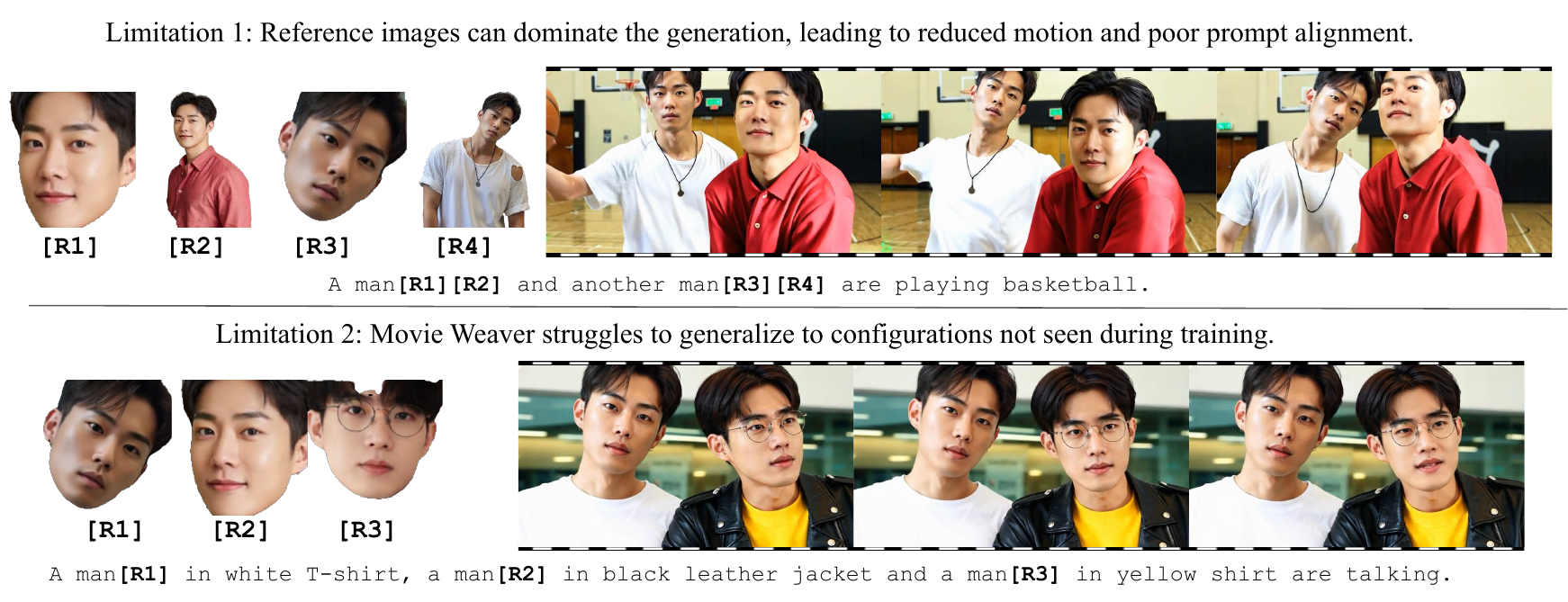}
    \vspace{-1em}
    \caption{\textbf{Limitations of Movie Weaver.} Reference images can dominate generation, resulting in 'big-face' videos. Our model also struggles to generalize to configurations not seen during training.}
    \label{fig:limitations}
    \vspace{-1em}
\end{figure*}

\subsection{Ablation study}
\label{sec:ablation_study}

\subsubsection{Anchored prompts and concept embeddings}
\label{sec:ablation_ap_ce}
In this section, we analyze the effects of the proposed Anchored Prompts (AP) and Concept Embeddings (CE). 
Identity blending was less apparent in ``face-body-animal" configurations because human animal embeddings are distinct. 
Thus, we conduct a human study on a two-face personalization evaluation dataset with 300 image pairs.
As in the top part of Figure~\ref{fig:ablation_ap_ce}, the baseline suffers from the identity blending issue, reflected by a low sep\_yes score, which indicates percentage of cases where the two generated faces are separable. 
Introducing concept embeddings, as seen in case (1), raises the sep\_yes score from 42.9\% to 98.2\%.
In case (2), anchored prompts further enhance identity preservation, increasing face1\_sim and face2\_sim scores. These scores represent the percentage of cases where a similar face to the left (face1) and right (face2) reference face is identifiable in the video.
Ablation with ``two-face-body" configuration can be found in  Appendix~\ref{appendix:ablation_twofacebody} and more details about human study can be found in Appendix~\ref{appendix:two_face_human_eval}.

When comparing CE with positional embeddings (PE), using the same setting as Figure~\ref{fig:ablation_ap_ce} Case (1), PE achieves 90.5\% sep\_yes score while CE achieves 98.2\%.

\subsubsection{Mixed training}
\label{sec:mix_training}

Our \workname is trained on various combinations of reference images, and this section examines the impact of this mixed training approach.
For a fair comparison, instead of initializing from a single-face personalization checkpoint, we start with a text-to-video base model.
As shown in Table~\ref{tab:mix_training}, we try different data configurations (1F: one-face, 2F: two-face and F-B-A: face-body-animal) and evaluate with a single-face personalization human study, which comprises 300 datapoints.
Compared to case (1), which uses only one-face data, mixed training with additional configurations improves both face\_sim (similarity of the generated character to the reference face) and face\_cons (consistency of the face throughout the video). 
We attribute this improvement to the increased data diversity provided by mixed training.

\begin{table}[t]
    \caption{\textbf{Ablation study of mixed training with multiple reference configurations.} 1F, 2F and F-B-A represents one-face, two-face and face-body-animal data, respectively. The metric face\_sim and face\_cons represents the similarity of character to the reference face and face consistency throughout the video.}
    \vspace{-0.5em}
    \centering
    \small
    \begin{tabular}{cccccc}
        \toprule
        \multirow{2}{*}{Case} & \multicolumn{3}{c}{Pretraining data} & \multicolumn{2}{c}{Human study metrics} \\
        \cmidrule(lr){2-4} \cmidrule(lr){5-6}
         & 1F & 2F & F-B-A & face\_sim$\uparrow$ & face\_cons$\uparrow$ \\
        \midrule
        1 & \checkmark &       &       & 13.1 & 70   \\
        2 & \checkmark & \checkmark &       & 29.5 & 81.3 \\
        3 & \checkmark & \checkmark & \checkmark & 46   & 91   \\
        \bottomrule
    \end{tabular}
    \vspace{-1em}
    \label{tab:mix_training}
\end{table}

\subsection{Limitations}

While \workname demonstrates strong multi-concept personalization capabilities, it has certain limitations. 
Firstly, the personalized videos often have limited overall motion compared to the results of base text-to-video model, and we sometimes see 'big-face' videos where faces occupy a large portion of the video frame.
We believe this occurs because the reference images can dominate the generation, leading to reduced motion and poor prompt alignment.
For instance, in the first example in Figure~\ref{fig:limitations}, we provide two half-body reference images with a prompt involving playing basketball.
The generated video reproduces the half-bodies but fails to align well with the action described in the prompt.
The underlying reason is \workname struggles to balance influence of reference images and text prompts when they are mismatched.
During training, videos of sports activities like basketball typically include full-body references, whereas during inference, users may provide less aligned inputs. Addressing the balance between reference images and prompts remains an area for future improvement.
Secondly, \workname struggles to generalize to configurations not seen during training.
In the second example in Figure~\ref{fig:limitations}, the goal is to generate a video with three people talking, but the result only features two people.
This is because we don't have any training videos that contain more than two people. 
Addressing this issue will require incorporating additional data configurations during pretraining, which we identify as a direction for future work.

%% file: sec/5_conclusion.tex
\section{Conclusion}

We present \workname to support tuning-free multi-concept video personalization. 
We alleviate the identity blending issue by explicitly associating the concept descriptions with reference images. 
With the pursuit of architectural simplicity and flexibility, we propose anchored prompts to inject unique tokens within text prompts and concept embeddings to encode the order of reference images.
Our results show that \workname can generate high quality personalized videos with diverse reference types, including face, body and animal images.

\section*{Acknowledgments}
This research was supported in part by ONR Minerva program, NSF CCF Grant No. 2107085, iMAGiNE - the Intelligent Machine Engineering Consortium at UT Austin, and a UT Cockrell School of Engineering Doctoral Fellowship.

%% file: sec/X_suppl.tex
\appendix

\section{Appendix}
\subsection{Data curation}

\paragraph{Video processing and filtering.}
For a given video, we uniformly sample five frames and apply a large-vocabulary object detector~\cite{zhou2022detecting} to each frame. The intersection of all detected objects across these frames is used to determine the objects present throughout the video. Using these detection results, we filter videos based on specific criteria. For example, to select videos featuring two people, we require two 'person' bounding boxes in the detection results. Similarly, for videos with one person and an animal, we ensure there is exactly one 'person' bounding box along with a 'dog' or 'cat' bounding box.

\paragraph{Two-face data curation.}
After obtaining the two-person video data, we utilize a suite of foundational models to generate anchored prompts and ordered reference images, as described in Section 4.3 of the main paper. Building on the approach of Movie Gen~\cite{moviegen}, we first employ the LLaMa3-Video~\cite{dubey2024llama} model to produce detailed text prompts for the video clips. These prompts follow a structured format, enabling the use of in-context learning to extract concept descriptions.
For example, given the input prompt: \texttt{Dentist Appointment. Senior woman smiling listening to her dentist during consultation.}, the outputs are two concept phrases: \texttt{[Senior woman smiling, her dentist]} and the anchored prompt: \texttt{Dentist Appointment. Senior woman smiling \textless{}ID1\textgreater{} listening to her dentist \textless{}ID2\textgreater{} during consultation.} Additional examples can be found in \href{https://jeff-liangf.github.io/projects/movieweaver/supp/in_context_twoface.txt}{\texttt{in\_context\_twoface.txt}}. Here, \texttt{\textless{}ID1\textgreater{}} and \texttt{\textless{}ID2\textgreater{}} represent \texttt{[R1]} and \texttt{[R2]}, respectively.
We further refine the output by ensuring that the concept phrases contain exactly two items and that both \texttt{\textless{}ID1\textgreater{}} and \texttt{\textless{}ID2\textgreater{}} appear in the anchored prompt.

\paragraph{Two-facebody data curation.}
After generating the two-face anchored prompt, creating the two-facebody prompt is straightforward. This involves replacing the original \texttt{\textless{}ID2\textgreater{}} with \texttt{\textless{}ID3\textgreater{} \textless{}ID4\textgreater{}} and \texttt{\textless{}ID1\textgreater{}} with \texttt{\textless{}ID2\textgreater{} \textless{}ID2\textgreater{}}. Additionally, we prepare the ordered two-face-body reference images to align with the updated prompt structure.

\paragraph{Face-body-animal data curation.}
We filter videos that feature one person with a pet (dog or cat). We use in-context examples to add three anchors to the original prompt. Examples can be found in \href{https://jeff-liangf.github.io/projects/movieweaver/supp/in_context_facebodyanimal.txt}{\texttt{in\_context\_facebodyanimal.txt}}

\subsection{Human evaluation}

\subsubsection{Two-face human evaluation}
\label{appendix:two_face_human_eval}

We conduct a human evaluation with 300 evaluation samples to ablate the effectiveness of the proposed anchored prompts and concept embeddings in Section 5.3.1. We provide the evaluation guidance as below. Besides the text guideline, we also include some visual examples to better help the annotators to judge.

\paragraph{Guidance.} This document describes how to do Movie Weaver two-face character consistency evaluation on generated video and their reference faces. 
The focus is on personalized video generation, where two reference faces are used to create a video, and the evaluation assesses how well the two generated characters maintain a consistent visual appearance compared to the two reference faces. We will be primarily focused on human characters (realistic or stylized).

\paragraph{Task description.} Annotators will be shown a set of two-faces and a generated video.
They are then asked to rate the character consistency level on the set of generated frames based on a few different questions related to the visual appearance of the person(s) in the reference image(s). 

\paragraph{Questions}
\begin{enumerate}[label=-]
    \item In the worst frame (they are not separable), are the two faces separable in the generated video (no fusion within two faces):
    \begin{itemize}
        \item [ ] 1 - Totally separable
        \item [ ] 2 - Somewhat separable
        \item [ ] 3 - Not separable
        \item [ ] 4 - Only one face or no face or more than two faces generated or visible
    \end{itemize}
    
    \textbf{Note:} In the specific example in Figure 3, annotators are expected to give the answer ``not separable''
    \item For the LEFT face in the reference, how well does the best aligned generated character’s face capture the person likeness? (Please first try the best to locate the best aligned character for the left reference face):
    \begin{itemize}
        \item [ ] 1 - Really similar
        \item [ ] 2 - Somewhat similar
        \item [ ] 3 - Not similar
        \item [ ] 4 - Only one face or no face or more than two faces generated or visible
    \end{itemize}
    
    \textbf{Note:} In this specific example in Figure 3, annotators are expected to give the answer ``Not similar''
    \item For the RIGHT face in the reference, how well does the best aligned generated character’s face capture the person likeness? (Please first try the best to locate the best aligned character for the right reference face):
    \begin{itemize}
        \item [ ] 1 - Really similar
        \item [ ] 2 - Somewhat similar
        \item [ ] 3 - Not similar
        \item [ ] 4 - Only one face or no face or more than two faces generated or visible
    \end{itemize}
    
    \textbf{Note:} In this specific example in Figure 3, annotators are expected to give the answer ``Not similar''
\end{enumerate}

\subsubsection{One-face human evaluation}

We perform a human evaluation with 300 samples to assess the effectiveness of mixed training, as discussed in Section 5.3.2. The evaluation protocol closely follows that of single-face personalized Movie Gen~\cite{moviegen}. Specifically, annotators are provided with a reference image and a generated video clip and asked to rate two aspects:
Face similarity (face\_sim): How well the generated character’s face matches the reference person in the best frame.
Face Consistency Score (face\_cons): How visually consistent the faces are across all frames containing the reference person.
Ratings are given on an absolute scale: “really similar,” “somewhat similar,” and “not similar” for identity, and “really consistent,” “somewhat consistent,” and “not consistent” for face consistency. Annotators are trained to adhere to specific labeling guidelines and are continuously audited to ensure quality and reliability.

\subsection{Additional results}

\subsubsection{Ablation on two-face-body configuration}
\label{appendix:ablation_twofacebody}
As shown in Table~\ref{tab:ablation_two_face_body}, ablation with ``two-face-body" showed similar trends to ``two-face" configurations. However, clothing details, like small logos in Figure~\ref{fig:teaser}, are harder to retain, likely due to the 256px reference resolution. Higher-resolution references may enhance clothing detail preservation.

\begin{table}[t]
    \centering
    \small
    \setlength{\tabcolsep}{3pt} 
    \vspace{-1em}
    \begin{tabular}{cccccc}
        \toprule
        \multirow{2}{*}{Case} & \multicolumn{2}{c}{Modules} & \multicolumn{3}{c}{Human study metrics} \\
        \cmidrule(lr){2-3} \cmidrule(lr){4-6}
        & AP & CE & sep\_yes$\uparrow$ & human1\_sim$\uparrow$ & human2\_sim$\uparrow$ \\
        \midrule
        Baseline &    &    & 54.8 & 12.3 & 16.5 \\
        (1)        &    & \checkmark & 98.8 & 66.7  & 69.4  \\
        (2)        & \checkmark & \checkmark & 98.0 & 72.3 & 71.1 \\
        \bottomrule
    \end{tabular}
    \caption{\textbf{Ablation study of Anchored Prompts (AP) and Concept Embeddings (CE) on ``two-face-body" config.}}
    \label{tab:ablation_two_face_body}
\end{table}

\subsubsection{Order of reference images}
\begin{figure}[t]
    \begin{center}
    \includegraphics[width=1.0\linewidth]{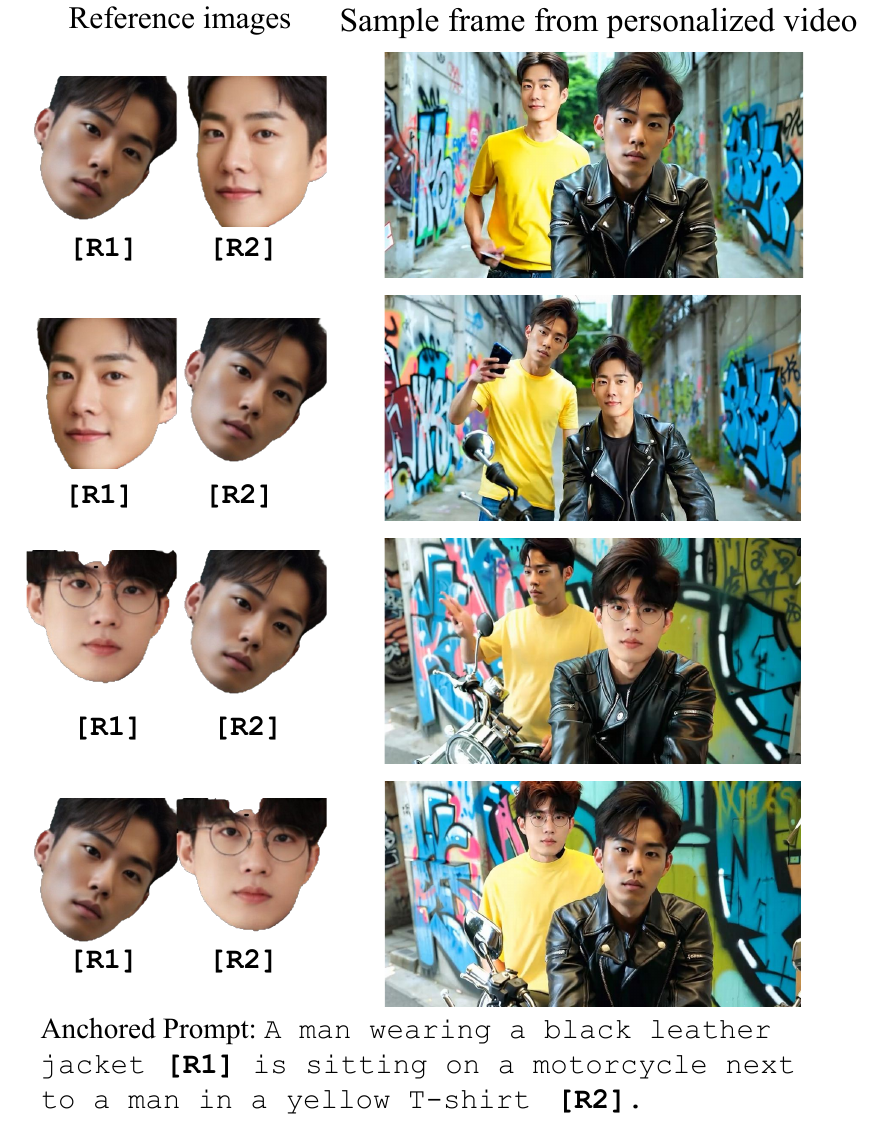}
    \end{center}
    \vspace{-1.5em}
        \caption{By changing the order of reference images, we can assign certain face to certain attributes.}
        \vspace{-1em}
    \label{fig:swap_reference}
\end{figure}

In this section, we examine how the order of reference images influences the final output. Since the order information is incorporated through concept embeddings, altering the sequence of reference images results in different videos, even with the same prompt. This effect is illustrated in Figure~\ref{fig:swap_reference}.

